\pdfoutput=1
\documentclass{article}

\PassOptionsToPackage{numbers,sort&compress}{natbib}
\usepackage[eandd, final]{neurips_2026}

\usepackage[utf8]{inputenc}
\usepackage[T1]{fontenc}
\usepackage{hyperref}
\hypersetup{pdftitle={Effect-Level Validation for Causal Discovery in Interactive Telemetry}, pdfkeywords={effect-level validation, causal discovery, sensitivity analysis, game telemetry}}
\usepackage{url}
\usepackage{booktabs}
\usepackage{placeins}

\usepackage{amsfonts}
\usepackage{microtype}
\usepackage{xcolor}

\usepackage{amsmath,amsthm,amssymb,bm}
\usepackage{tabularx}
\usepackage{graphicx}
\usepackage{subcaption}
\usepackage{enumitem}
\usepackage{tikz}
\usetikzlibrary{shapes,arrows,calc,arrows.meta,positioning}
\usepackage{float}

\newtheorem{definition}{Definition}[section]

\title{Effect-Level Validation for Causal Discovery in Interactive Telemetry}
\author{
  Hoang Dang \\
  Independent Researcher \\
  Vietnam \\
  \texttt{hoangdang112023@gmail.com} \\
  \And
  Luan Pham \\
  RMIT University \\
  Australia \\
  \texttt{luan.pham@rmit.edu.au} \\
  \And
  Minh Nguyen \\
  Florida Atlantic University \\
  USA \\
  \texttt{minhnguyen@fau.edu}
}

\begin{document}

\maketitle

\begin{abstract}
Causal discovery is increasingly used to analyze telemetry from interactive systems, but a plausible graph does not guarantee an identifiable or reliable product decision. We study this problem for one query: whether early competitive gameplay increases Day-1 retention in a deployed Role-playing game (RPG). We propose an admissibility-first framework that treats discovered graphs as structural hypotheses, keeps only graphs that identify the target effect and have adequate treatment-control overlap, and validates admissible effects through cross-algorithm stability, placebo, subsampling, and E-value diagnostics. On real telemetry, under minimal constraints the admissible runs that return Directed Acyclic Graphs (DAGs) are those whose graphs leave player-versus-player (PvP) with no parents, so they estimate the confounded raw difference; under the full constraint set all are admissible. Runs returning Partial Ancestral Graphs (PAGs) are not evaluated by the DAG gate and are reported separately. The admissible DAGs agree on one adjustment set and return a positive, cohort-conditional risk-difference average treatment effect (ATE), down from a raw retention gap; the estimate keeps its sign under placebo and subsampling checks, but its magnitude changes with the adjustment set and the constraint specification. A known-truth simulation shows that the checks screen identification and stability given the measured variables and cannot certify the absence of unmeasured confounding, and a second game-telemetry dataset shows that identifiability rates do not transfer. Graph-level recovery is therefore a weak proxy for causal reliability; discovery pipelines for decision support should be evaluated at the query and effect level.
\end{abstract}

\section{Introduction}
\label{sec:intro}

Game teams rarely need a causal graph for its own sake.
They need a trustworthy answer to a concrete decision question: does early exposure to competitive gameplay increase next-day retention?
If the answer is yes, it provides evidence for investing in (PvP) onboarding, one lever among several the team may consider.
If the answer is a selection artifact, high-engagement players seek out PvP and would have retained regardless, then the same investment is wasted.
Getting this wrong translates directly into misallocated product effort and false confidence in a design decision~\citep{kang2024match, kim2026analyzing}.

The standard approach is causal discovery: learn a causal graph from observational telemetry, then use it to identify and estimate the effect.
Most evaluation frameworks assess whether the discovered graph resembles a ground-truth structure, using metrics such as Structural Hamming Distance, edge precision, or recall on synthetic benchmarks~\citep{chakraborty2023causil, raghu2018evaluation}.
This conflates two distinct questions.
A graph can score well on structural recovery yet fail to support point identification of the target estimand, because critical edge orientations remain unresolved, or because the graph encodes latent structure that prevents adjustment-based identification.
Conversely, structurally divergent graphs can yield the same valid effect estimate when the target causal pathway is preserved through alternative routes.
In feedback-driven telemetry, behavioral metrics aggregate across time, and engagement patterns create spurious associations, making this structural-to-causal reliability gap severe~\citep{banerjee2020large}.

We argue that the central evaluation question is not \emph{which graph looks most correct} but \emph{which graphs support a stable and defensible answer to the target query}.
Section~\ref{sec:method} formalizes this distinction as an admissibility gate.
We therefore propose an \textbf{admissibility-first, effect-level validation framework} that treats each discovered graph as a candidate explanation and asks whether it can identify the target effect, not whether it resembles an unavailable ground-truth graph.
A graph passes the admissibility screen if it identifies the target effect and has adequate treatment-control overlap.
Graphs that fail this screen are excluded from effect estimation regardless of statistical fit.
For passing graphs, reliability is assessed through cross-algorithm stability, refutation tests, and E-value analysis.

Our running example is a deployed Web3 RPG.
The target query is whether early PvP engagement in the first 24 hours causally increases Day-1 retention (hours 24--48).
Candidate covariates include platform commitment (\texttt{Web3} wallet connection), onboarding friction (\texttt{Time\_Play\_Level1}), baseline progression (\texttt{Total\_PvE\_Battle},player-versus-environment battles), and session intensity (\texttt{Total\_Session}).
The game setting makes structural uncertainty especially visible: players self-select into optional PvP features, telemetry aggregates conflate causes with consequences, and the live-service environment evolves through long-term feedback.
We use ``domain-general'' only for the workflow (specify the target query, derive constraints from measurement timing and the data schema, test identifiability per query, report constraint sensitivity), not for a fixed constraint set or for the empirical conclusions that follow from it.

Applying this framework to real game telemetry, we find that under minimal constraints the admissible DAG-returning runs are those whose graphs leave PvP with no parents and so report the confounded raw difference, while under the full constraint set PC, GRaSP and BOSS return structurally different graphs that imply the same adjustment set and the same estimate, about 14--17 percentage points depending on whether overlap trimming is applied.
This is agreement on one estimand, not independent recovery of the effect, and the estimate changes with the adjustment set and with the constraint specification (Section~\ref{sec:results}).
It keeps its sign under placebo and subsampling checks, but a known-truth simulation shows that such checks cannot certify the absence of unmeasured confounding.
Graph-level metrics are therefore a weak screen for query-level causal reliability.

\noindent Our key contributions are:
\begin{enumerate}[leftmargin=*, nolistsep, topsep=4pt]
    \item \textbf{Admissibility-first, effect-centric framework.}
    We propose a pipeline that treats discovered graphs as hypotheses and filters them by admissibility (identifiability and positivity) before effect estimation, shifting from graph recovery to query reliability under structural uncertainty.

    \item \textbf{Effect-level validation protocol.}
    We introduce a validation protocol combining overlap diagnostics, refutation tests (placebo and subsampling), and sensitivity analysis (E-values) to assess when a discovered causal effect can be trusted.

    \item \textbf{Evidence on what cross-algorithm agreement does and does not show.}
    Constraint-based (PC) and score-based (GRaSP and BOSS) searches return structurally different graphs (pairwise structural Hamming distance 1--4) that imply the same adjustment set and hence the same estimate, for the reason given above (Section~\ref{sec:method}, Section~\ref{sec:results}).

    \item \textbf{Characterization of identifiability bottlenecks.}
    Whether a discovery output admits a point-identified query depends on the constraint set and the cohort (all DAG runs under the full set on the telemetry, 0--67\% on the three WoWAH cohorts); identifiability and stability cannot be inferred from algorithm families or graph-level metrics alone, and on a second dataset both the rates and the constraints that matter differ.
\end{enumerate}

\noindent Our replication package, including code and data, is available at \url{https://github.com/dangvanconghoang/NeurIPS_Effect-Level-Validation-for-Causal-Discovery}.
 \section{Related work}

Causal inference is increasingly used in telemetry-rich online systems because observational behavior reflects confounding, selective exposure, and policy feedback. This is visible in recommender systems and marketing optimization, where predictive associations are often insufficient for intervention decisions~\citep{gao2024causal, xu2025causal, zhou2024decision, banerjee2020large}. Most work in this line starts from a specified treatment, design, or causal structure. Our setting adds a different question: when the graph itself is learned from telemetry, which discovered structures can still support the target effect?

Causal discovery has also been used for operational telemetry, especially root-cause analysis in distributed and microservice systems~\citep{ikram2022root, chakraborty2023causil, pham2024root}. These studies show that learned causal graphs can be useful under partial observability, but their goal is localization or ranking, not effect identification for a target treatment. We instead ask whether a learned structure identifies a user-level treatment effect and remains stable enough for decision support.

Game telemetry makes this problem especially sharp. Player behavior and system state co-evolve, optional features induce self-selection, and retention outcomes mix product effects with pre-existing engagement. Prior work documents retention drivers and confounding in games, including simultaneity in competitive play, matchmaking effects, progression and churn models, and mediation pathways involving community and toxicity~\citep{christiansen2019resolving, kang2024match, park2017achievement, karmakar2022improved, liu2025sympathy}. These studies motivate causal reasoning in games, but they do not evaluate effects derived from discovery-produced graphs.

Evaluation of causal discovery remains difficult when no ground-truth graph is available. Graph-recovery metrics and synthetic benchmarks can miss whether a learned structure supports the intervention query of interest~\citep{raghu2018evaluation, gentzel2019case, faller2024self}. Structural Intervention Distance moves evaluation toward intervention semantics, but still requires a reference DAG or CPDAG and does not directly cover partial ancestral graph (PAG) outputs from latent-aware methods~\citep{peters2015structural}. Related evaluation work also shows that benchmark scores, assumptions, and hyperparameters can change conclusions about causal methods~\citep{curth2021really, montagna2023assumption}. Our framework follows this query-level view: it treats discovered graphs as hypotheses, filters them by admissibility, and validates the resulting effect through cross-algorithm stability checks and refutation tests.
 \section{Method}
\label{sec:method}

Two graphs can share the same skeleton and score similarly under graph-recovery metrics, yet differ in whether the target query is identifiable, as Figure~\ref{fig:main_core_idea} illustrates. The framework below treats this distinction as a first-order evaluation criterion rather than a minor structural detail.

\begin{figure}[t]
\centering
\begin{tikzpicture}[
  node distance=1.2cm and 1.6cm,
  box/.style={draw, rounded corners, minimum width=1.0cm, minimum height=0.55cm, font=\small},
  lbl/.style={draw=none, font=\small\itshape},
  >=Stealth
]
\node[box] (WA) {$W$};
\node[box, right=of WA] (TA) {$T$};
\node[box, right=of TA] (YA) {$Y$};
\draw[->] (WA) -- (TA);
\draw[->] (WA) to[bend left=28] (YA);
\draw[->] (TA) -- (YA);
\node[lbl, below=0.55cm of TA, xshift=-0.85cm]
  {\textbf{Graph A} --- effect identifiable};
\begin{scope}[xshift=6.8cm]
\node[box] (WB) {$W$};
\node[box, right=of WB] (TB) {$T$};
\node[box, right=of TB] (YB) {$Y$};
\draw[->] (WB) -- (TB);
\draw[->] (WB) to[bend left=28] (YB);
\draw[->] (TB) -- (YB);
\draw[<->] (TB) to[bend right=28] (YB);
\node[lbl, below=0.55cm of TB, xshift=-0.85cm]
  {\textbf{Graph B} --- effect \emph{not} identifiable};
\end{scope}
\end{tikzpicture}
\caption{Two graphs with the same skeleton yet opposite implications for the query $T \to Y$. Graph A: $W$ is an observed confounder; backdoor adjustment identifies $\mathbb{E}[Y \mid do(T)]$. Graph B: $T \to Y$ coexists with a latent common cause $T \leftrightarrow Y$; adjusting for $W$ cannot block the hidden path $T \leftarrow H \to Y$, so no observed adjustment set point-identifies the effect.}
\label{fig:main_core_idea}
\end{figure}

\subsection{Setup}

The framework takes observational data, a treatment--outcome pair with a target effect, and constraints from time ordering and variable construction. It asks, across multiple discovered graphs, whether the target effect is identifiable and stable, not which single graph is best.

In our running example, \(T\) is early PvP exposure (first 24 hours) and \(Y\) is Day-1 retention (hours 24--48). Covariates include \texttt{Web3}, \texttt{Time\_Play\_Level1}, \texttt{Total\_PvE\_Battle}, and \texttt{Total\_Session}. As in Section~\ref{sec:intro}, ``domain-general'' here refers only to the workflow, not to the constraint set; this setting makes structural uncertainty visible throughout the analysis. The pipeline below operationalizes this question in five stages, moving from candidate graphs discovered without reference to the target query to a reliability verdict on the one estimand we care about.

\subsection{Pipeline}

The framework proceeds in five steps:

\begin{enumerate}[leftmargin=*, nolistsep, topsep=4pt]
    \item \textbf{Discover candidate graphs.} Run multiple causal discovery algorithms on the observational telemetry. We evaluate constraint-based (PC, FCI, CFCI, FCI-MAX), score-based (BOSS, GRaSP), and hybrid methods (GFCI, BOSS-FCI, GRaSP-FCI). Algorithmic details and graph deviation metrics are in Appendix~\ref{app:algorithms} and Table~\ref{tab:graph_metrics}.
    \item \textbf{Enforce domain constraints.} Apply \(\mathcal{C}\) to each graph, forbidding impossible edges.
    \item \textbf{Screen for admissibility.} Check whether each constrained graph admits a valid identification strategy for the target estimand and satisfies positivity. Graphs that fail either criterion are excluded from effect estimation and recorded as identification failures; PAG outputs, which the DAG backdoor gate does not evaluate, are recorded separately as not evaluated (Section~\ref{sec:admissibility}).
    \item \textbf{Estimate the target effect.} For admissible graphs, estimate the ATE via backdoor adjustment with a doubly robust estimator; regression adjustment and inverse probability weighting (IPW) serve as secondary references. The estimator's cross-fitting folds and nuisance models are explicitly seeded for reproducibility (Section~\ref{sec:effect-est}).
    \item \textbf{Assess reliability.} Evaluate effect stability across algorithms and significance thresholds, and apply placebo, subsampling, and E-value sensitivity tests.
\end{enumerate}

Discovery is performed on temporally sliced variables, assuming acyclicity within each time window. The target query is defined within explicitly ordered windows; later-measured variables never cause earlier-measured ones, and how fully that applies to the treatment specifically is a declared choice (below), not a uniform default.

\subsection{Domain constraints}

Domain knowledge is encoded as a set of minimal forbidden and required edge directions. Temporal constraints prevent future outcomes from causing earlier behaviors. Semantic constraints reflect how telemetry variables are constructed: platform-level attributes (\texttt{Web3}) are not caused by downstream gameplay events. These constraints are not intended to enforce any specific causal pathway between treatment and outcome, but to exclude causally meaningless configurations.

Constraints come from three evidence sources, ordered by how contestable each is: (i) \emph{temporal closure}: an edge from a later-measured to an earlier-measured variable is forbidden, with landmarks following the same rule; (ii) \emph{measurement prerequisites}: the part--whole relation of composite aggregates is encoded from the feature definition; and (iii) \emph{system preconditions}: an edge is constrained only where one variable is a technical precondition for another to exist. For sensitivity analysis, these sources are partitioned into four removable implementation groups (temporal, platform, measurement, and semantic); the ``semantic'' group contains the single timing-supported PvP$\nrightarrow$onboarding exclusion and is not a fourth evidence source. The main constraint set C3 implements temporal closure fully for the outcome but only for that treatment--onboarding pair, keeping treatment-side knowledge minimal. The stricter C4 set forbids PvP$\to$every pre-treatment covariate and therefore implements treatment-side temporal closure fully (Section~\ref{sec:adjset}). Leave-one-group-out analyses are reported for the main telemetry cohort and the 336-hour replication cohort (Section~\ref{sec:loo}).

\subsection{Admissibility: identification and positivity}
\label{sec:admissibility}

\begin{definition}[Admissible graph]
A discovered graph \(G\) is \emph{admissible} for query \((T, Y)\) if it satisfies: (i) all pre-specified temporal and semantic constraints, (ii) a valid identification strategy for the estimand, and (iii) adequate overlap for the adjustment set in the data.
\end{definition}

Graphs that fail this criterion are excluded from downstream analysis regardless of their statistical fit. Enforcing domain constraints can alter edge orientations or leave adjacencies unresolved: structures may satisfy conditional independence tests yet admit no valid adjustment set, identification failures are informative outcomes of the pipeline, not errors.

Beyond graphical identification, each admissible adjustment set must satisfy positivity. We assess overlap via propensity score diagnostics and trim observations outside the common support region. Graphs that admit a valid adjustment set but violate overlap after trimming are flagged as inadmissible. For low-prevalence treatments we additionally check that estimates lie within their logical bounds and that the effective sample size is adequate; runs that fail are reported as implausible, not as estimates.

PAG outputs represent uncertainty from latent confounding. The DAG backdoor gate does not evaluate them, so we label them \emph{not evaluated under the DAG backdoor gate} rather than as failures; this is a limitation of the evaluation rule, not a deficiency of FCI-family methods. Their PAG-calculus estimates~\citep{jaber2022causal} form a separate reporting track (Appendix~\ref{app:pag}; Tables~\ref{tab:pag_causal_discovery_methods}--\ref{tab:pag_effects}); all main comparisons use DAG-returning discovery methods only. Table~\ref{tab:claim_assumption} (Appendix~\ref{app:claims}) maps each claim to its assumptions, the evidence provided, and the failure condition for that claim.

\subsection{Effect estimation}
\label{sec:effect-est}

We use DoWhy~\citep{dowhy, JMLR:v25:22-1258} for effect estimation after the admissibility gate. Effects are identified via the backdoor criterion and estimated with a doubly robust (DR) estimator, which provides protection against misspecification of the outcome or treatment model. Regression adjustment and IPW are reported as secondary references. The estimator's cross-fitting folds and nuisance models are seeded; before seeding, repeating an identical task moved the estimate by up to 1.9 percentage points over 30 seeds (across-seed SD 0.0036). Effects are on the risk-difference scale. When common-support trimming is applied, the estimand is the ATE over the overlap-satisfying subpopulation.

\subsection{Reliability checks}

We assess effect reliability through four complementary diagnostics:

\begin{enumerate}[leftmargin=*, nolistsep, topsep=4pt]
    \item \textbf{Cross-algorithm stability.} Effects across admissible graphs are compared for sign, magnitude, and interval estimates. Agreement across structurally different graphs shows that the estimate does not depend on which admissible graph is returned; it is agreement on one adjustment set, not independent recovery of the effect, because divergent algorithms can share a common bias.
    \item \textbf{Placebo treatment test.} The true treatment is replaced with a randomly permuted variable. A non-null placebo effect indicates residual confounding or model misspecification.
    \item \textbf{Data subset test.} Effects are re-estimated on random subsamples; high sensitivity suggests overfitting or sample-specific artifacts rather than a true effect.
    \item \textbf{Sensitivity to unmeasured confounding.} E-values~\citep{vanderweele2017sensitivity} quantify how strong an unmeasured confounder must associate with treatment and outcome to explain away the observed effect. On the real telemetry we report them as a sensitivity summary, not a pass/fail test: they bound how strong confounding would have to be and cannot establish that none exists. The known-truth simulation (Section~\ref{sec:knowntruth}) also thresholds the E-value into a \emph{cleared every check} indicator, but only to aggregate runs for reporting, not to claim confounding below it is ruled out.
\end{enumerate}
 \section{Data}
\label{sec:data}

\subsection{Dataset overview}

We use behavioral telemetry from a deployed Web3 role-playing game (RPG) collected over 5 months. The raw telemetry contains 77,698 users and 26 telemetry-derived features captured through Firebase, covering sessions, progression milestones, economic transactions, and system interactions. The game combines a mandatory PvE progression path with an optional Arena/PvP mode, making early PvP adoption a natural self-selected treatment rather than a mandatory progression step.

The analysis cohort is constructed in three steps. First, we restrict attention to paying players, yielding 1,376 users. Second, quality filters and missingness removal produce the complete-case analytic sample used for discovery and estimation (\(n=1{,}280\)). Third, overlap trimming for the final effect estimates leaves 1,170 users in the common-support sample.

The unit of analysis is the individual user. For each user, the treatment is first PvP exposure, which can occur only after the eligibility time $t_0$ defined below. All covariates are constructed from events occurring strictly before $t_0$, hence before any PvP exposure, ensuring no post-treatment information is used for discovery, adjustment, or overlap diagnostics.

Data is structured into two non-overlapping windows: a covariate window (hours 0--24) aggregating all pre-treatment behavioral features, and an outcome window (hours 24--48) observing Day-1 retention. We restrict analysis to users active until PvP eligibility time $t_0$; covariates are computed strictly from data prior to $t_0$, and users who churn before $t_0$ are excluded from both groups. Because first PvP exposure is defined after $t_0$, eligibility is a baseline criterion for this estimand rather than conditioning on a post-treatment variable. It does narrow the population substantially, so the estimate does not generalize to users who churn before eligibility; in the replication of Section~\ref{sec:wowah}, only 11,795 of 58,655 newly created avatars (20.1\%) reach eligibility.

\subsection{Descriptive statistics and self-selection evidence}

The six variables are defined in Table~\ref{tab:variables} (Appendix~\ref{app:variables}): treatment (PvP), outcome (R1), and four covariates covering platform commitment, friction, progression, and engagement. Table~\ref{tab:descriptive} (Appendix~\ref{app:variables}) summarizes the treatment split: \(263/1280 = 20.55\%\) engaged in PvP ($T=1$) within the first day, while \(1017/1280 = 79.45\%\) did not ($T=0$). The overall Day-1 retention rate (R1) is 70.39\%. PvP-exposed users show 89\% retention versus 65\% for non-exposed (raw gap: 24 pp), but they are more active across all covariates, Web3 connection (73\% vs.\ 42\%), PvE battles (18.87 vs.\ 8.67), sessions (5.32 vs.\ 3.50). This severe self-selection requires adjustment-based estimation.
 \section{Results}
\label{sec:results}

Under minimal constraints, admissible DAG-returning runs leave PvP with no parents and report only the confounded raw difference at C0, and none are admissible at C1; under the full constraint set all are admissible, with an effect of 14--17 percentage points that keeps its sign under placebo and subsampling checks but is conditional on the adjustment set and constraint specification (Section~\ref{sec:adjset}); a known-truth simulation (Section~\ref{sec:knowntruth}) shows what the checks cannot certify.

\subsection{Target estimand and identification}

Identification depends on the declared constraints. At C0, 10 of 21 DAG-returning runs are admissible, all through graphs in which PvP has no parents, so they identify the unadjusted risk difference (0.239), which the descriptive statistics of Section~\ref{sec:data} show to be confounded; at C1 none is; at C3, all 21 are (PC, GRaSP and BOSS at all seven thresholds). PAG outputs are not evaluated by the DAG backdoor gate and are reported separately. The target estimand is the risk-difference ATE, \(\tau = \mathbb{E}[Y \mid do(T=1)] - \mathbb{E}[Y \mid do(T=0)]\), where \(T\) is PvP exposure in the first 24 hours and \(Y\) is Day-1 retention during hours 24--48. Identification is attempted via the backdoor criterion for each discovered structure after constraint enforcement. DoWhy identifies \texttt{Web3} as the minimal sufficient adjustment set for the graphs admissible at C2 and C3; other covariates are retained in the structural model but excluded from adjustment. All reported effects are cohort-conditional ATEs for paying users active until \(t_0\) (1,280 of 77,698 users); any onboarding decision would require external validation in the intended population.

\textbf{Why only \texttt{Web3}.} The set was not hand-selected: in the admissible DAGs, \texttt{Total\_PvE\_Battle} and \texttt{Total\_Session} are oriented as descendants of PvP, so DoWhy excludes them, and \texttt{Time\_Play\_Level1} is not adjacent to PvP and lies on no backdoor path. Since every covariate is aggregated strictly before first PvP exposure (Section~\ref{sec:data}), that orientation is a discovery output, not a claim of this paper, and judged against the data design it is wrong; C3 does not forbid it because it deliberately keeps injected knowledge minimal. Forbidding it (C4) leaves the target estimand unchanged but changes the adjustment set and the estimate (Section~\ref{sec:adjset}).

\subsection{Main causal effect}

Adjustment reduces the raw retention gap (24 pp on the full sample, 19 pp on the common-support sample) to a 14--15 pp cohort-conditional effect on the common-support sample ($n=1{,}170$, seeded estimators). Among admissible graphs, early PvP exposure increases Day-1 retention with \(\hat{\tau}_{\text{REG}} = 0.155\), \(\hat{\tau}_{\text{IPW}} = 0.146\), and \(\hat{\tau}_{\text{DR}} = 0.139\). The DR estimator is primary; its 95\% confidence interval is $(0.081,\,0.198)$. All three estimators agree in sign and magnitude. On the full sample ($n=1{,}280$) the same seeded estimator, and every admissible discovery run, gives 0.174; the difference is attributable entirely to overlap trimming.

\subsection{Constraint ladder ablation}
\label{sec:constraint-ladder}

Table~\ref{tab:constraint-ladder} shows that, on our telemetry, identifiability is not monotone in the constraint level and is not the same as correctness. At C0, 10 of 21 runs are admissible only because their graphs leave PvP with no parents, so they return the confounded raw difference 0.239; at C1 none is; at C2 and C3, all 21 are, with the same estimate (0.174). On WoWAH the pattern differs by cohort and is not monotone on the 336-hour cohort (Table~\ref{tab:wowah}), so we do not claim a general ordering.

We evaluate four constraint levels from weakest to strongest: \textbf{C0} forbids only \texttt{R1}$\to$\texttt{other}; \textbf{C1} additionally forbids \texttt{other}$\to$\texttt{Web3}; \textbf{C2} requires the measurement edge \texttt{Total\_PvE\_Battle}$\to$\texttt{Total\_Session}; \textbf{C3}, the full set used elsewhere, also forbids \texttt{PvP}$\to$\texttt{Time\_Play\_Level1} (semantic group). Each level is evaluated across a grid of seven independence thresholds $\alpha$. A run is one (algorithm, $\alpha$) tuple. We aggregate two metrics: \emph{identifiable rate} (fraction of runs where $\widehat{\text{ATE}}$ is defined) and \emph{effect stability} ($\operatorname{SD}$ of defined ATEs).

\begin{table}[!htbp]
\centering
\caption{Constraint ladder ablation on the telemetry (DAG-returning algorithms PC, GRaSP and BOSS; 7 thresholds each, 21 runs per level; full sample, $n=1{,}280$; seeded estimator). The C0 runs return the unadjusted risk difference. SD is zero because every admissible run at a level implies the same adjustment set. C0 is one run; GRaSP varies over repeats (Appendix~\ref{app:responsible_reporting}).}
\label{tab:constraint-ladder}
\begin{tabular}{@{}p{1.8cm}p{2.5cm}p{2.3cm}p{1.6cm}@{}}
\toprule
Constraint & Identifiable rate (\%) & Mean ATE (RD) & SD (ATE) \\
\midrule
C0 & 47.62 & 0.239 & 0.000 \\
C1 & 0     & NaN   & NaN   \\
C2 & 100   & 0.174 & 0.000 \\
C3 & 100   & 0.174 & 0.000 \\
\bottomrule
\end{tabular}
\end{table}

\subsection{Sensitivity to the adjustment set and constraint specification}
\label{sec:adjset}

The main analysis adjusts for \texttt{Web3} alone. Table~\ref{tab:adjset} tests what happens if that choice is wrong, on the common-support sample: the DR estimate falls from 0.139 under \{\texttt{Web3}\} to 0.075 when \texttt{Total\_PvE\_Battle} is added, to 0.043 under the three parents that C4 implies, and to 0.056 under all four covariates, where every interval includes zero; REG and IPW move the same way. Cross-fitted AIPW on all four covariates gives 0.050 (SE 0.040) with logistic and 0.027 (SE 0.210) with gradient-boosted nuisance models, neither different from zero.

\begin{table}[!htbp]
\centering
\caption{Sensitivity of the PvP effect to the adjustment set (common-support sample, $n=1{,}170$; seeded). REG: regression adjustment; IPW: inverse-probability weighting; DR: doubly robust (DRLearner) with 95\% CI. \{\texttt{Web3}\} is the minimal set returned by DoWhy under C3; \{\texttt{Web3}, \texttt{Total\_PvE\_Battle}, \texttt{Total\_Session}\} is the parent set of PvP in every completed C4 run.}
\label{tab:adjset}
\begin{tabular}{@{}p{5.4cm}cccc@{}}
\toprule
Adjustment set & REG & IPW & DR & 95\% CI (DR) \\
\midrule
None (raw difference) & 0.194 & --- & --- & --- \\
\{\texttt{Web3}\} (main analysis) & 0.155 & 0.146 & 0.139 & [0.081, 0.198] \\
\{\texttt{Web3}, \texttt{Total\_PvE\_Battle}\} & 0.064 & 0.082 & 0.075 & [$-0.009$, 0.158] \\
\{\texttt{Web3}, \texttt{Total\_PvE\_Battle}, \texttt{Total\_Session}\} (C4-motivated sensitivity) & 0.044 & 0.053 & 0.043 & [$-0.047$, 0.133] \\
All four pre-treatment covariates & 0.035 & 0.049 & 0.056 & [$-0.033$, 0.144] \\
\bottomrule
\end{tabular}
\end{table}

C3 applies the temporal rule to the outcome but not to the treatment, although every covariate is measured before the treatment landmark. Applying it to both is equally defensible a priori (C4: C3 plus PvP$\to$covariate forbidden). In the runs all 21 C4 DAG runs return $\text{parents}(\text{PvP}) = \{\texttt{Web3}, \texttt{Total\_PvE\_Battle}, \texttt{Total\_Session}\}$, but the DAG gate rejects every one: the same graphs keep an undirected \texttt{Time\_Play\_Level1}--\texttt{Total\_PvE\_Battle} edge, off the treatment's parent set, and the gate rejects any graph containing an undirected edge, so the pipeline returns no C4 estimate. GRaSP varies here (this parent set in 121 of 140 repeated draws), and none produced an estimate. Adjusting for that parent set in a separately specified sensitivity graph gives 0.043 (third row); this is not a pipeline-produced C4 estimate. The \{\texttt{Web3}\} estimate must therefore be read as conditional on C3, and the parent-set sensitivity shows how strongly the result depends on that particular constraint specification.

\subsection{Leave-one-constraint-out}
\label{sec:loo}

Dropping one of the four constraint groups at a time (Table~\ref{tab:loo}, Appendix~\ref{app:loo}) shows that identifiability depends strongly on the constraints. Estimates agree among variants that return the same adjustment set; dropping the platform constraint on telemetry instead yields graphs in which \texttt{Web3} is not a parent of PvP (10 of 21 runs) and the confounded raw difference 0.239 rather than 0.174. The semantic constraint is inert on both datasets. The measurement constraint is necessary on telemetry (0 of 21 without it) but costly on WoWAH, where dropping it raises graph identification from 14/21 to 17/21 and within-bounds from 7/21 to 14/21. Dropping the platform constraint leaves 1/21 WoWAH runs graph-identified and none within bounds; dropping the temporal constraint leaves 7/21 graph-identified and none within bounds. BOSS runs that reach the upstream iteration cap are retained in the denominator as nonconverged and contribute no estimate. GRaSP varies in the platform-dropped variant (20 repeats: 33.3--52.4\%), not in the measurement-dropped one. What transfers is the procedure for deriving and stress-testing constraints, not a fixed constraint set.

\subsection{Overlap and positivity diagnostics}

Figure~\ref{fig:overlap} (Appendix~\ref{app:positivity}) shows imperfect but adequate overlap after trimming 8.6\% of users; detailed balance diagnostics are reported in Table~\ref{tab:positivity_diagnostics}. Common support leaves 1,170 of 1,280 users. On the full sample the maximum absolute standardized mean difference (SMD) falls from 1.16 to 0.10 after inverse probability weighting (IPW; Table~\ref{tab:positivity_diagnostics}) and the IPW effective sample size is 836; after trimming they are 0.104 and 742.

\subsection{Refutation and sensitivity analysis}

\textbf{Placebo treatment.} Replacing the true treatment with a random permutation yields an effect near zero ($-0.004$, $p=0.96$), indicating the observed result is not an artifact of the adjustment model.

\textbf{Data subset stability.} Re-estimating on random subsamples (10\%, 50\%, 80\%, 90\%; mean of 100 draws, seeded) yields mean estimates of 0.176, 0.106, 0.133 and 0.131 with no sign reversal; they do not reproduce the full-sample value (0.139) exactly, so the magnitude is sample-sensitive.

\textbf{Sensitivity to unmeasured confounding.} Poisson regression with robust standard errors gives risk ratio \(RR = 1.19\) (95\% CI: 1.12--1.26). The E-value is 1.66 (CI bound: 1.48): an unmeasured confounder would need $RR \geq 1.66$ with both treatment and outcome to explain away the observed effect. This is not strong evidence that unmeasured confounding is absent, and moderate unmeasured confounding could overturn the estimate: for context, three of the four observed covariates have risk-ratio associations of 2.9--8.7 with PvP and about 1.5 with retention (high versus low, median split for counts; \texttt{Time\_Play\_Level1}: 1.1 and 1.0), so a hidden confounder as strong as those would approach the E-value, and we do not treat it as implausible. In the known-truth simulation below, 96.7\% of runs in a latent-confounding regime cleared the placebo, subsampling and E-value checks together while remaining wrong by more than 5 percentage points; these checks quantify how strong such confounding would have to be, and cannot rule it out.

None of these checks reverses the sign of the effect, which is sign-stable but magnitude-sensitive, and they cannot eliminate all bias. Under the minimal constraints, identification is either absent (C1) or obtained only through graphs in which PvP has no parents (C0, 10 of 21 runs), which return the confounded raw difference 0.239; full results are in Appendix~\ref{app:minimal_constraints}.

\subsection{Run-level outcome taxonomy}
\label{sec:taxonomy}

Table~\ref{tab:taxonomy} (Appendix~\ref{app:taxonomy}) classifies the 21 DAG-returning runs at the full constraint set (C3): all are admissible and convergent, and none fail positivity. At weaker constraints the dominant outcome is identification failure: all 21 runs at C1 and 11 of 21 at C0 return a graph with an undirected or partially oriented edge, which the gate treats as not identified, and the other 10 runs at C0 are admissible only through graphs in which PvP has no parents. The six PAG-returning algorithms (42 runs) are outside this denominator (Appendix~\ref{app:pag}).

\subsection{Effect stability across discovery algorithms}

Among DAG-returning methods, PC, GRaSP and BOSS are admissible at every threshold under C3 and return the same estimate ($\hat{\tau} = 0.174$ on the full sample, variance $0$). PAG-returning methods are not evaluated by the DAG gate and are reported separately in Appendix~\ref{app:pag}.

At the structural level, PC, GRaSP and BOSS orient the \texttt{Web3}$\to$\texttt{PvP} edge and maintain a path from \texttt{PvP} to \texttt{R1} sufficient for backdoor identification. FCI methods return graphs in which all 42 telemetry runs at C3 carry a circle or bidirected mark (e.g.\ \texttt{o-o} on \texttt{PvP}--\texttt{Web3}), which the backdoor gate treats as not point-identified; this reflects the scope of the DAG gate, not an error of these methods.

\subsection{What the cross-algorithm agreement shows}
\label{sec:graphcomparison}

On our telemetry the admissible runs at C2 and C3 (21 PC, GRaSP and BOSS runs, counted at both levels) span three structurally distinct graphs, with pairwise structural Hamming distance 1--4, and all return an identical estimate (Figure~\ref{fig:admissible_graphs}, Table~\ref{tab:graph_comparison}, Appendix~\ref{app:graphcomp}). All three imply the same parent set for the treatment, \{\texttt{Web3}\}: they differ only where the estimand does not depend on them, so convergence is agreement on one estimand, not independent recovery of the effect, confirming the caution of Section~\ref{sec:method}. The 10 admissible C0 runs add two graphs in which PvP has no parents, a second adjustment set and a different estimate. On WoWAH the admissible graphs imply two (336h) and three (168h) adjustment sets, and the in-bounds estimates agree only within a set.

\subsection{Known-truth evaluation}
\label{sec:knowntruth}

The claim that passing all checks means the effect is correct cannot be tested on real telemetry, so we built four regimes over a temporally ordered structural causal model ($n=1{,}280$, matching the empirical cohort) with the interventional ATE exact by construction: 4 regimes $\times$ 100 replications $\times$ 4 algorithms $\times$ 3 thresholds $=$ 4,800 discovery runs, 900 DAG runs per regime. A run is \emph{safe} if its estimate is within 5 percentage points (pp) of the truth.

\begin{table}[!htbp]
\centering
\caption{Known-truth evaluation (DAG-returning runs; Appendix~\ref{app:knowntruth}). ``Wrong despite every check'': among the runs that cleared the gate and the placebo, subsampling and E-value checks together, the share that were nevertheless not safe. The gate rejects any unresolved edge, as on the telemetry.}
\label{tab:knowntruth}
\setlength{\tabcolsep}{4pt}
\begin{tabularx}{\columnwidth}{@{}>{\raggedright\arraybackslash}p{2.7cm}cccc>{\centering\arraybackslash}X@{}}
\toprule
Regime & True ATE & Gate pass & Safe & Mean abs. error & Wrong despite every check \\
\midrule
Identifiable & 0.101 & 90.1\% & 85.4\% & 1.8 pp & 10.0\% \\
Latent confounder & 0.102 & 99.6\% & 0\% & 13.9 pp & 100\% \\
Orientation ambiguous & 0.116 & 98.3\% & 1.3\% & 11.1 pp & 98.6\% \\
Null (no causal path) & 0.000 & 2.6\% & --- & --- & 0 runs cleared all checks \\
\bottomrule
\end{tabularx}
\end{table}

Where identification holds the framework behaves as intended: 90.1\% of runs are admitted and 85.4\% are safe, and in the null regime 877 of 900 runs are rejected and none of the 23 survivors clears the full reliability suite. Under latent confounding it does not: the gate admits 99.6\% of runs, 96.7\% also clear placebo, subsampling and E-value together, and every one of those is wrong, by 13.9 pp on average. A sweep over $n \in \{500, 1280, 5000\}$ shows this is structural: the share clearing every check rises from 76\% to 98\% while the mean error stays near 14 pp. The framework screens identification and stability given the measured variables, and cannot certify the absence of unmeasured confounding. SHD correlates with absolute ATE error when pooled across regimes ($+0.466$) but does not separate safe from unsafe answers (51.1\% of low-SHD runs are still wrong by more than 5 pp), so graph similarity is a weak screen for decision usefulness rather than an uninformative one (Appendix~\ref{app:knowntruth}).

\subsection{Replication on a second dataset}
\label{sec:wowah}

We replicate the pipeline on the World of Warcraft Avatar History crawl~\citep{lee2011world} (36.5M records, 91,064 avatars) through the identical discovery $\to$ identification $\to$ estimation workflow. Each column keeps its role from the main analysis, and the PvP-eligibility landmark is derived from the observed progression data rather than fixed a priori (Appendix~\ref{app:wowah}). This is an independent replication within game telemetry; other domains remain future work.

\begin{table}[!htbp]
\centering
\caption{DAG-returning runs (PC, GRaSP and BOSS; 7 thresholds each, 21 runs per cell, seeded) on WoWAH. Each cell reports the percentage producing a defined effect / the percentage also lying within the logical bounds $[-1,1]$ of a risk difference. BOSS runs hitting the upstream iteration cap are retained as nonconverged, contributing no estimate.}
\label{tab:wowah}
\begin{tabular}{@{}lcccc@{}}
\toprule
Cohort & C0 & C1 & C2 & C3 \\
\midrule
WoWAH 24h ($n=9{,}592$) & 0/0\% & 0/0\% & 0/0\% & 0/0\% \\
WoWAH 168h ($n=9{,}436$) & 19.0/9.5\% & 33.3/23.8\% & 61.9/33.3\% & 61.9/33.3\% \\
WoWAH 336h ($n=9{,}338$) & 4.8/0\% & 81.0/66.7\% & 66.7/33.3\% & 66.7/33.3\% \\
\bottomrule
\end{tabular}
\end{table}

The operational claim carries over: a discovered graph frequently cannot answer the query, and on the 24-hour cohort no DAG run at any constraint level produces an estimand. The rates differ by cohort and are not generally monotone: graph identification rises through C3 on the 168-hour cohort, whereas on the 336-hour cohort it peaks at C1 and falls from 81.0\% to 66.7\% at C2--C3. The within-bounds rates are lower still, so a graph-defined estimand is not by itself a usable estimate.
 \section{Conclusion}

Causal discovery is useful for telemetry only when discovered structures can answer the decision query. In our game telemetry case, under minimal constraints the admissible DAG-returning runs are those whose graphs leave PvP with no parents and so estimate the confounded raw difference, and under the full constraint set all are admissible; PAG outputs are not evaluated by the DAG gate (Section~\ref{sec:admissibility}) and are reported separately. Among admissible DAGs, PC, GRaSP and BOSS imply the same adjustment set and yield the same cohort-conditional effect of early PvP exposure on Day-1 retention, about 14--17 percentage points; the estimate keeps its sign under placebo and subsampling checks, but its magnitude changes with the adjustment set and the constraint specification, and the checks cannot certify the absence of unmeasured confounding.

This shifts the evaluation target: graph-recovery metrics ask whether a learned graph resembles a reference structure, whereas decision support asks whether the target effect is identifiable and stable. Treating discovered graphs as structural hypotheses and validating effects across algorithms offers a practical standard where no ground-truth graph is available.

\paragraph{Limitations.}
Four limits: the evidence is one game and one treatment--outcome pair, with the second dataset replicating within game telemetry and other domains left to future work; the ATE covers only paying users active until $t_0$; PAG outputs are not evaluated under the DAG gate (Section~\ref{sec:admissibility}), a rule limit rather than a method flaw; and constraints and time windows need re-specifying elsewhere.
 
\section*{Acknowledgments}
We thank the anonymous reviewers for their careful and constructive feedback.

\bibliographystyle{plainnat}
\bibliography{references}

\begin{thebibliography}{32}
\providecommand{\natexlab}[1]{#1}
\providecommand{\url}[1]{\texttt{#1}}
\expandafter\ifx\csname urlstyle\endcsname\relax
  \providecommand{\doi}[1]{doi: #1}\else
  \providecommand{\doi}{doi: \begingroup \urlstyle{rm}\Url}\fi

\bibitem[Andrews et~al.(2019)Andrews, Ramsey, and Cooper]{andrews2019learning}
Bryan Andrews, Joseph Ramsey, and Gregory~F Cooper.
\newblock Learning high-dimensional directed acyclic graphs with mixed
  data-types.
\newblock In \emph{The 2019 ACM SIGKDD Workshop on Causal Discovery}, pages
  4--21. PMLR, 2019.

\bibitem[Andrews et~al.(2023)Andrews, Ramsey, Sanchez~Romero, Camchong, and
  Kummerfeld]{andrews2023fast}
Bryan Andrews, Joseph Ramsey, Ruben Sanchez~Romero, Jazmin Camchong, and Erich
  Kummerfeld.
\newblock Fast scalable and accurate discovery of dags using the best order
  score search and grow shrink trees.
\newblock \emph{Advances in neural information processing systems},
  36:\penalty0 63945--63956, 2023.

\bibitem[Banerjee et~al.(2020)Banerjee, Mukherjee, Dutta, and
  Ghosh]{banerjee2020large}
Trambak Banerjee, Gourab Mukherjee, Shantanu Dutta, and Pulak Ghosh.
\newblock A large-scale constrained joint modeling approach for predicting user
  activity, engagement, and churn with application to freemium mobile games.
\newblock \emph{Journal of the American Statistical Association}, 2020.

\bibitem[Bl{{\"o}}baum et~al.(2024)Bl{{\"o}}baum, G{{\"o}}tz, Budhathoki,
  Mastakouri, and Janzing]{JMLR:v25:22-1258}
Patrick Bl{{\"o}}baum, Peter G{{\"o}}tz, Kailash Budhathoki, Atalanti~A.
  Mastakouri, and Dominik Janzing.
\newblock Dowhy-gcm: An extension of dowhy for causal inference in graphical
  causal models.
\newblock \emph{Journal of Machine Learning Research}, 25\penalty0
  (147):\penalty0 1--7, 2024.
\newblock URL \url{http://jmlr.org/papers/v25/22-1258.html}.

\bibitem[Chakraborty et~al.(2023)Chakraborty, Garg, Agarwal, Chauhan, and
  Saini]{chakraborty2023causil}
Sarthak Chakraborty, Shaddy Garg, Shubham Agarwal, Ayush Chauhan, and
  Shiv~Kumar Saini.
\newblock Causil: Causal graph for instance level microservice data.
\newblock In \emph{Proceedings of the ACM Web Conference 2023}, pages
  2905--2915, 2023.

\bibitem[Christiansen et~al.(2019)Christiansen, Gensby, and
  Weber]{christiansen2019resolving}
Anders~Harboell Christiansen, Emil Gensby, and Bryan~S Weber.
\newblock Resolving simultaneity bias: using features to estimate causal
  effects in competitive games.
\newblock In \emph{2019 IEEE Conference on Games (CoG)}, pages 1--8. IEEE,
  2019.

\bibitem[Curth et~al.(2021)Curth, Svensson, Weatherall, and Van
  Der~Schaar]{curth2021really}
Alicia Curth, David Svensson, Jim Weatherall, and Mihaela Van Der~Schaar.
\newblock Really doing great at estimating cate? a critical look at ml
  benchmarking practices in treatment effect estimation.
\newblock In \emph{Thirty-fifth conference on neural information processing
  systems datasets and benchmarks track (round 2)}, 2021.

\bibitem[Faller et~al.(2024)Faller, Vankadara, Mastakouri, Locatello, and
  Janzing]{faller2024self}
Philipp~M Faller, Leena~C Vankadara, Atalanti~A Mastakouri, Francesco
  Locatello, and Dominik Janzing.
\newblock Self-compatibility: Evaluating causal discovery without ground truth.
\newblock In \emph{International Conference on Artificial Intelligence and
  Statistics}, pages 4132--4140. PMLR, 2024.

\bibitem[Gao et~al.(2024)Gao, Zheng, Wang, Feng, He, and Li]{gao2024causal}
Chen Gao, Yu~Zheng, Wenjie Wang, Fuli Feng, Xiangnan He, and Yong Li.
\newblock Causal inference in recommender systems: A survey and future
  directions.
\newblock \emph{ACM Transactions on Information Systems}, 42\penalty0
  (4):\penalty0 1--32, 2024.

\bibitem[Gentzel et~al.(2019)Gentzel, Garant, and Jensen]{gentzel2019case}
Amanda Gentzel, Dan Garant, and David Jensen.
\newblock The case for evaluating causal models using interventional measures
  and empirical data.
\newblock \emph{Advances in Neural Information Processing Systems}, 32, 2019.

\bibitem[Ikram et~al.(2022)Ikram, Chakraborty, Mitra, Saini, Bagchi, and
  Kocaoglu]{ikram2022root}
Azam Ikram, Sarthak Chakraborty, Subrata Mitra, Shiv Saini, Saurabh Bagchi, and
  Murat Kocaoglu.
\newblock Root cause analysis of failures in microservices through causal
  discovery.
\newblock \emph{Advances in Neural Information Processing Systems},
  35:\penalty0 31158--31170, 2022.

\bibitem[Jaber et~al.(2022)Jaber, Ribeiro, Zhang, and
  Bareinboim]{jaber2022causal}
Amin Jaber, Adele Ribeiro, Jiji Zhang, and Elias Bareinboim.
\newblock Causal identification under markov equivalence: Calculus, algorithm,
  and completeness.
\newblock \emph{Advances in Neural Information Processing Systems},
  35:\penalty0 3679--3690, 2022.

\bibitem[Kang et~al.(2024)Kang, Suh, and Kim]{kang2024match}
Hyunjae Kang, Changwoo Suh, and Huy~Kang Kim.
\newblock Match experiences affect interest: Impacts of matchmaking and
  performance on churn in a competitive game.
\newblock \emph{Heliyon}, 10\penalty0 (3), 2024.

\bibitem[Karmakar et~al.(2022)Karmakar, Liu, Mukherjee, Che, and
  Dutta]{karmakar2022improved}
Bikram Karmakar, Peng Liu, Gourab Mukherjee, Hai Che, and Shantanu Dutta.
\newblock Improved retention analysis in freemium role-playing games by jointly
  modelling players' motivation, progression and churn.
\newblock \emph{Journal of the Royal Statistical Society Series A: Statistics
  in Society}, 185\penalty0 (1):\penalty0 102--133, 2022.

\bibitem[Kim et~al.(2026)Kim, Kim, Kim, and Li]{kim2026analyzing}
Jongho Kim, Kihan Kim, Hansol Kim, and Shanshan Li.
\newblock Analyzing player churn in esports games: a survival analysis of
  league of legends log data.
\newblock \emph{Sport, Business and Management: An International Journal},
  16\penalty0 (4):\penalty0 453--474, 2026.

\bibitem[Lam et~al.(2022)Lam, Andrews, and Ramsey]{lam2022greedy}
Wai-Yin Lam, Bryan Andrews, and Joseph Ramsey.
\newblock Greedy relaxations of the sparsest permutation algorithm.
\newblock In \emph{Uncertainty in Artificial Intelligence}, pages 1052--1062.
  PMLR, 2022.

\bibitem[Lee et~al.(2011)Lee, Chen, Cheng, and Lei]{lee2011world}
Yeng-Ting Lee, Kuan-Ta Chen, Yun-Maw Cheng, and Chin-Laung Lei.
\newblock World of warcraft avatar history dataset.
\newblock In \emph{Proceedings of the second annual ACM conference on
  Multimedia systems}, pages 123--128, 2011.

\bibitem[Liu et~al.(2025)Liu, Tang, and Williams]{liu2025sympathy}
Mingxuan Liu, Jack~Lipei Tang, and Dmitri Williams.
\newblock Sympathy for the devil: Serial mediation models for toxicity,
  community, and retention.
\newblock \emph{Media and Communication}, 13, 2025.

\bibitem[Montagna et~al.(2023)Montagna, Mastakouri, Eulig, Noceti, Rosasco,
  Janzing, Aragam, and Locatello]{montagna2023assumption}
Francesco Montagna, Atalanti Mastakouri, Elias Eulig, Nicoletta Noceti, Lorenzo
  Rosasco, Dominik Janzing, Bryon Aragam, and Francesco Locatello.
\newblock Assumption violations in causal discovery and the robustness of score
  matching.
\newblock \emph{Advances in Neural Information Processing Systems},
  36:\penalty0 47339--47378, 2023.

\bibitem[Ogarrio et~al.(2016)Ogarrio, Spirtes, and Ramsey]{ogarrio2016hybrid}
Juan~Miguel Ogarrio, Peter Spirtes, and Joe Ramsey.
\newblock A hybrid causal search algorithm for latent variable models.
\newblock In \emph{Conference on probabilistic graphical models}, pages
  368--379. PMLR, 2016.

\bibitem[Park et~al.(2017)Park, Cha, Kwak, and Chen]{park2017achievement}
Kunwoo Park, Meeyoung Cha, Haewoon Kwak, and Kuan-Ta Chen.
\newblock Achievement and friends: Key factors of player retention vary across
  player levels in online multiplayer games.
\newblock In \emph{Proceedings of the 26th international conference on world
  wide web companion}, pages 445--453, 2017.

\bibitem[Peters and B{\"u}hlmann(2015)]{peters2015structural}
Jonas Peters and Peter B{\"u}hlmann.
\newblock Structural intervention distance for evaluating causal graphs.
\newblock \emph{Neural computation}, 27\penalty0 (3):\penalty0 771--799, 2015.

\bibitem[Pham et~al.(2024)Pham, Ha, and Zhang]{pham2024root}
Luan Pham, Huong Ha, and Hongyu Zhang.
\newblock Root cause analysis for microservice system based on causal
  inference: How far are we?
\newblock In \emph{Proceedings of the 39th IEEE/ACM International Conference on
  Automated Software Engineering}, pages 706--715, 2024.

\bibitem[Raghu et~al.(2018{\natexlab{a}})Raghu, Poon, and
  Benos]{raghu2018evaluation}
Vineet~K Raghu, Allen Poon, and Panayiotis~V Benos.
\newblock Evaluation of causal structure learning methods on mixed data types.
\newblock In \emph{Proceedings of 2018 ACM SIGKDD Workshop on Causal
  Discovery}, pages 48--65. PMLR, 2018{\natexlab{a}}.

\bibitem[Raghu et~al.(2018{\natexlab{b}})Raghu, Ramsey, Morris, Manatakis,
  Sprites, Chrysanthis, Glymour, and Benos]{raghu2018comparison}
Vineet~K Raghu, Joseph~D Ramsey, Alison Morris, Dimitrios~V Manatakis, Peter
  Sprites, Panos~K Chrysanthis, Clark Glymour, and Panayiotis~V Benos.
\newblock Comparison of strategies for scalable causal discovery of latent
  variable models from mixed data.
\newblock \emph{International journal of data science and analytics},
  6\penalty0 (1):\penalty0 33--45, 2018{\natexlab{b}}.

\bibitem[Ramsey et~al.(2012)Ramsey, Zhang, and Spirtes]{ramsey2012adjacency}
Joseph Ramsey, Jiji Zhang, and Peter~L Spirtes.
\newblock Adjacency-faithfulness and conservative causal inference.
\newblock \emph{arXiv preprint arXiv:1206.6843}, 2012.

\bibitem[Ramsey et~al.(2025)Ramsey, Andrews, and Spirtes]{ramsey2025efficient}
Joseph Ramsey, Bryan Andrews, and Peter Spirtes.
\newblock Efficient latent variable causal discovery: Combining score search
  and targeted testing.
\newblock \emph{arXiv preprint arXiv:2510.04263}, 2025.

\bibitem[Sharma and Kiciman(2020)]{dowhy}
Amit Sharma and Emre Kiciman.
\newblock Dowhy: An end-to-end library for causal inference.
\newblock \emph{arXiv preprint arXiv:2011.04216}, 2020.

\bibitem[Spirtes et~al.(2001)Spirtes, Glymour, and
  Scheines]{spirtes2001causation}
Peter Spirtes, Clark Glymour, and Richard Scheines.
\newblock \emph{Causation, prediction, and search}.
\newblock The MIT press, 2001.

\bibitem[VanderWeele and Ding(2017)]{vanderweele2017sensitivity}
Tyler~J VanderWeele and Peng Ding.
\newblock Sensitivity analysis in observational research: introducing the
  e-value.
\newblock \emph{Annals of internal medicine}, 167\penalty0 (4):\penalty0
  268--274, 2017.

\bibitem[Xu et~al.(2025)Xu, Ji, Li, Ge, Tan, and Zhang]{xu2025causal}
Shuyuan Xu, Jianchao Ji, Yunqi Li, Yingqiang Ge, Juntao Tan, and Yongfeng
  Zhang.
\newblock Causal inference for recommendation: Foundations, methods, and
  applications.
\newblock \emph{ACM Transactions on Intelligent Systems and Technology},
  16\penalty0 (3):\penalty0 1--51, 2025.

\bibitem[Zhou et~al.(2024)Zhou, Huang, Li, Jiang, Zheng, Cheng, and
  Lin]{zhou2024decision}
Hao Zhou, Rongxiao Huang, Shaoming Li, Guibin Jiang, Jiaqi Zheng, Bing Cheng,
  and Wei Lin.
\newblock Decision focused causal learning for direct counterfactual marketing
  optimization.
\newblock In \emph{Proceedings of the 30th ACM SIGKDD Conference on Knowledge
  Discovery and Data Mining}, pages 6368--6379, 2024.

\end{thebibliography}

\appendix
\section{How causal models answer interventional queries}
\label{app:primer}

A causal graph is a set of assumptions about which variables can influence which. To ask what Day-1 retention would be if every player were exposed to PvP early, we must separate the effect of PvP from the effect of everything that makes players choose PvP. \emph{Identification} asks whether, under the graph's assumptions, that interventional quantity can be computed from observed data at all. \emph{Backdoor adjustment} is one way to do so: if the graph lets us find a set of measured variables that blocks every non-causal route from treatment to outcome, then comparing exposed and unexposed players who share the same values of those variables recovers the causal effect. If instead the graph leaves the orientation between treatment and outcome unresolved, or marks a hidden common cause between them, no such set may exist and the query is not identified, however well the graph fits the data.

A DAG commits to one orientation for every edge. A PAG leaves some orientations open to express uncertainty about hidden causes, so a PAG and a DAG can give different answers to the same query: the DAG answers it under the orientation it chose, whereas the PAG answers only what holds under every orientation it allows, which may be nothing. This is why the admissibility gate asks, graph by graph, whether the query can be answered, and why an estimate is read as conditional on the graph and the adjustment set that produced it.

\section{Exploratory PAG-based point estimates}
\label{app:pag}

This separate track holds the six PAG-returning algorithms (FCI, CFCI, FCI-MAX, GFCI, BOSS-FCI, GRaSP-FCI; 42 runs per constraint level), which the DAG backdoor gate does not evaluate. We scored them with the PAG-calculus estimator~\citep{jaber2022causal} on the common-support sample under the C1 constraint set (script \texttt{experiment/replication/pag\_track.py}; per-run output in \texttt{pag\_track\_estimates.csv}). The estimator returns a point estimate for only 5 of the 42 runs, all from CFCI, and those five range from 0.00 to 0.39. Where the PAG-calculus identification fails the query is not identified from the PAG, which is consistent with, but not the same as, the DAG-gate rejection. No main-text number depends on this track, and the estimates are not comparable with the DAG results.

\begin{table}[t]
    \centering
    \caption{Summary of causal discovery method families. PAG outputs are separated from the DAG-based main analysis because partial orientation can imply a different target effect or identification rule. The key point is that methods producing valid causal outputs need not contribute to the same stability claim.}
    \label{tab:pag_causal_discovery_methods}
    \begin{tabular}{@{} p{3.5cm} p{2.5cm} p{2cm} @{}}
        \toprule
        Method family                  & Typical output          &  Used in main stability claims \\
        \midrule
        PC / GRaSP / BOSS       & DAG                   & Yes                              \\
        \addlinespace
        FCI, CFCI, FCI-MAX, GFCI, BOSS-FCI, GRaSP-FCI & PAG                                   & No                     \\
        \bottomrule
    \end{tabular}
\end{table}

\begin{table}[t]
\centering
\caption{PAG-calculus estimates for the six PAG-returning algorithms (seven significance levels each; common-support sample; C1 constraints; identical across two runs). A run is counted as defined when the PAG-calculus estimator returns a point estimate.}
\label{tab:pag_effects}
\begin{tabular}{lcc}
\toprule
Method & Runs with an estimate (of 7) & Range of estimates \\
\midrule
FCI & 0 & --- \\
GFCI & 0 & --- \\
BOSS-FCI & 0 & --- \\
GRaSP-FCI & 0 & --- \\
FCI-MAX & 0 & --- \\
CFCI & 5 & 0.00--0.39 \\
\bottomrule
\end{tabular}
\end{table}

\section{Variable definitions}
Below we define the treatment, outcome, and covariates used in the causal analysis, and report descriptive statistics by treatment status. See Table~~\ref{tab:variables}
\label{app:variables}

\begin{table}[h]
    \centering

    \caption{Variables used in the causal analysis: treatment, outcome, and four pre-treatment covariates covering platform commitment, onboarding friction, baseline progression, and engagement intensity.}
    \label{tab:variables}
    \begin{tabular}{@{}llp{4.2cm}@{}}
    \toprule
    Variable & Type & Definition \\
    \midrule
    R1 (Outcome) & Binary & Day-1 retention: 1 if user returned during hours 24--48, 0 otherwise \\
    PvP (Treatment) & Binary & Early PvP exposure: participation in competitive mode during first 24 hours \\
    Web3 & Binary & Wallet connection status (proxy for platform commitment) \\
    Time\_Play\_Level1 & Continuous & Time spent on Level 1 (onboarding friction proxy) \\
    Total\_PvE\_Battle & Integer & Number of PvE battles (baseline progression) \\
    Total\_Session & Integer & Total login sessions (engagement intensity) \\
    \bottomrule
    \end{tabular}
\end{table}

\begin{table}[t]
    \centering
    \caption{Descriptive statistics by treatment status. The raw 24 pp retention gap is heavily confounded by self-selection: PvP-exposed users differ strongly on Web3 status, PvE progression, and session count.}
    \label{tab:descriptive}
    \begin{tabular}{@{}lccc@{}}
    \toprule
    Variable & Overall & PvP=1 & PvP=0 \\
    \midrule
    R1 (Retention) & 0.70 & \textbf{0.89} & 0.65 \\
    Web3 (Wallet) & 0.48 & \textbf{0.73} & 0.42 \\
    Time\_Play\_Level1 & 7.83 & \textbf{8.07} & 7.77 \\
    Total\_PvE\_Battle & 10.76 & \textbf{18.87} & 8.67 \\
    Total\_Session & 3.87 & \textbf{5.32} & 3.50 \\
    \bottomrule
    \end{tabular}
\end{table}

\section{Effect behavior under minimally constrained discovery}
\label{app:minimal_constraints}

Under the minimal constraints C0 (only R1$\to$other forbidden) and C1 (additionally other$\to$Web3 forbidden), the DAG gate admits 10 of 21 runs at C0 and none at C1. The 10 admissible C0 runs (PC at all seven thresholds, GRaSP at three) return fully directed graphs in which PvP has no parents, so the identified estimand is the unadjusted risk difference, 0.239, the confounded raw gap of Table~\ref{tab:descriptive}; they are admissible because the graph identifies an effect, not because it is the right one. The other 11 runs at C0 and all 21 at C1 return graphs with an undirected or partially oriented edge, which the gate treats as not identified. The leave-one-constraint-out analysis (Table~\ref{tab:loo}) shows which constraint groups are active on each dataset.

\section{Algorithm details}
\label{app:algorithms}

We evaluate nine causal discovery algorithms grouped by inference mechanism:

Constraint-based: PC, FCI~\citep{spirtes2001causation}, CFCI~\citep{ramsey2012adjacency}, and FCI-MAX~\citep{raghu2018comparison} use conditional independence (CI) tests to prune edges. They employ the Degenerate Gaussian Likelihood Ratio Test~\citep{andrews2019learning} to handle mixed data.

Score-based: BOSS~\citep{andrews2023fast} and GRaSP~\citep{lam2022greedy} explore permutation space and maximize the Degenerate Gaussian BIC score. The Tetrad build used for BOSS (Appendix~\ref{app:responsible_reporting}) places an iteration cap on it; a run that reaches it is recorded as nonconverged, retained in the denominator, and contributes no estimate.

Hybrid: GFCI~\citep{ogarrio2016hybrid}, BOSS-FCI, and GRaSP-FCI~\citep{ramsey2025efficient} combine score-based search with FCI-style refinement.

\begin{table}[t]
    \centering
    \caption{Graph deviation metrics of unconstrained discovery (no domain knowledge, default significance level) relative to the nine-edge domain-consistent reference graph (Figure~\ref{fig:causal_graph}), using the six analysis variables. Precision, recall, tp, fp and fn count arrowheads (CFCI returns no arrowhead, so its precision is undefined); adj\_* count adjacencies. GRASP is stochastic and shown as the mean of 20 runs (SHD 7--9); the other columns were identical in every run. Metrics are descriptive: low SHD does not imply effect reliability, and high SHD does not preclude admissible effect estimation.}
    \label{tab:graph_metrics}
    \begin{tabular}{lcccc}
        \toprule
        \multicolumn{5}{l}{Panel A: PC, GFCI, FCI, and BOSS-FCI} \\
        \midrule
        Metric & PC & GFCI & FCI & BOSS-FCI \\
        \midrule
        precision & 0.30 & 0.36 & 0.50 & 0.42 \\
        recall & 0.33 & 0.44 & 0.78 & 0.56 \\
        tp & 3.00 & 4.00 & 7.00 & 5.00 \\
        fp & 7.00 & 7.00 & 7.00 & 7.00 \\
        fn & 6.00 & 5.00 & 2.00 & 4.00 \\
        adj\_precision & 0.80 & 0.80 & 0.80 & 0.80 \\
        adj\_recall & 0.89 & 0.89 & 0.89 & 0.89 \\
        adj\_tp & 8.00 & 8.00 & 8.00 & 8.00 \\
        adj\_fp & 2.00 & 2.00 & 2.00 & 2.00 \\
        adj\_fn & 1.00 & 1.00 & 1.00 & 1.00 \\
        shd & 8.00 & 9.00 & 8.00 & 9.00 \\
        \bottomrule
    \end{tabular}

    \medskip

    \begin{tabular}{lccccc}
        \toprule
        \multicolumn{6}{l}{Panel B: GRASP-FCI, GRASP, BOSS, FCI-MAX, and CFCI} \\
        \midrule
        Metric & GRASP-FCI & GRASP & BOSS & FCI-MAX & CFCI \\
        \midrule
        precision & 0.42 & 0.43 & 0.57 & 0.46 & --- \\
        recall & 0.56 & 0.37 & 0.44 & 0.67 & 0.00 \\
        tp & 5.00 & 3.35 & 4.00 & 6.00 & 0.00 \\
        fp & 7.00 & 4.65 & 3.00 & 7.00 & 0.00 \\
        fn & 4.00 & 5.65 & 5.00 & 3.00 & 9.00 \\
        adj\_precision & 0.80 & 0.72 & 0.67 & 0.80 & 0.80 \\
        adj\_recall & 0.89 & 0.89 & 0.89 & 0.89 & 0.89 \\
        adj\_tp & 8.00 & 8.00 & 8.00 & 8.00 & 8.00 \\
        adj\_fp & 2.00 & 3.15 & 4.00 & 2.00 & 2.00 \\
        adj\_fn & 1.00 & 1.00 & 1.00 & 1.00 & 1.00 \\
        shd & 9.00 & 8.80 & 9.00 & 8.00 & 11.00 \\
        \bottomrule
    \end{tabular}
\end{table}

\begin{figure}[t]
    \centering
    \includegraphics[width=\linewidth]{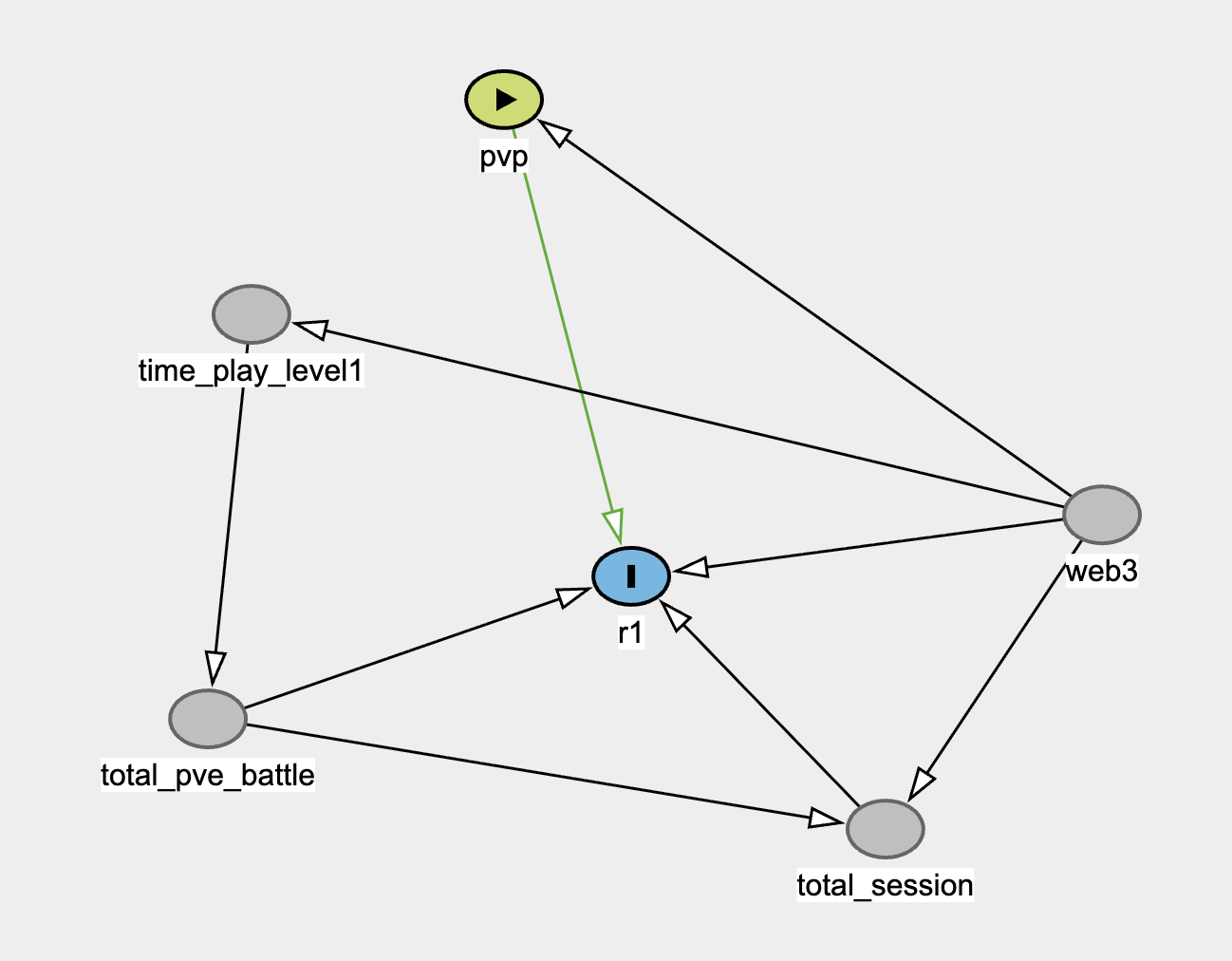}
    \caption{Reference graph from domain constraints. Not ground truth, one admissible hypothesis supporting backdoor identification of the target estimand.}
    \label{fig:causal_graph}
\end{figure}

\section{Edge constraints specification}
\label{app:constraints}

All constraints are pre-specified based on telemetry construction and system design, applied prior to discovery, and do not encode assumptions about effect magnitude or sign. The semantic constraint is inert on both datasets (Table~\ref{tab:loo}); we keep it in the specification and report it as inert rather than load-bearing.

\section{Model validation details}
\label{app:validation}

\subsection{Functional causal model assumptions}

We assess FCM assumptions using independence tests between noise terms and parents. Across all mechanisms (LogReg, Decision Tree, HistGradientBoosting), p-values equal 1.0 for Total\_Session, Time\_Play\_Level1 and Total\_PvE\_Battle under all three mechanism classes, indicating no evidence against assumed causal mechanisms. Calibration and retention-variance diagnostics are summarized in Tables~\ref{tab:crps} and~\ref{tab:icc}.

\subsection{Calibration metrics}

\begin{table}[t]
    \centering
    \caption{Calibration performance (CRPS) for each modeled node under three mechanism classes. Lower values indicate better calibration. The key pattern is that retention and PvP are calibrated well across mechanisms, while the continuous telemetry variables are only fair, with logistic regression performing best overall for the retention node. The Assessment column is a qualitative label per node, not a formal threshold on CRPS.}
    \label{tab:crps}
    \begin{tabular}{lcccc}
        \toprule
        Node & LogReg & DT & HistGB & Assessment \\
        \midrule
        R1 & 0.163 & 0.223 & 0.180 & Very Good \\
        PvP & 0.166 & 0.169 & 0.170 & Very Good \\
        Time\_Play\_Level1 & 0.502 & 0.509 & 0.501 & Fair \\
        Total\_PvE\_Battle & 0.528 & 0.527 & 0.523 & Fair \\
        Total\_Session & 0.477 & 0.476 & 0.474 & Fair \\
        \bottomrule
    \end{tabular}
\end{table}

Logistic Regression achieves lowest CRPS for the retention outcome (0.163), indicating superior calibration.

\subsection{Intrinsic causal contribution}

\begin{table}[t]
    \centering
    \caption{Intrinsic causal contribution (ICC) to the variance of retention, by mechanism class (seeded). ``Intrinsic'' is the share attributed to R1's own noise term and ``Parents'' the share attributed to its modeled parents (Web3, PvP, Time\_Play\_Level1, Total\_PvE\_Battle, Total\_Session); the two sum to 100. The decision-tree mechanism attributes the most variance to the parents, and logistic regression the least.}
    \label{tab:icc}
    \begin{tabular}{lcc}
        \toprule
        Mechanism & Intrinsic (\%) & Parents (\%) \\
        \midrule
        Logistic Regression & 61.6 & 38.4 \\
        Decision Tree & 37.8 & 62.2 \\
        HistGradientBoosting & 58.5 & 41.5 \\
        \bottomrule
    \end{tabular}
\end{table}

Under Logistic Regression about 62\% of retention variance is intrinsic to R1 and about 38\% is attributed to its modeled parents; the attribution depends on the mechanism class (Table~\ref{tab:icc}).

\section{Positivity diagnostics}
\label{app:positivity}

\begin{figure}[htbp]
    \centering
    \includegraphics[width=0.9\linewidth]{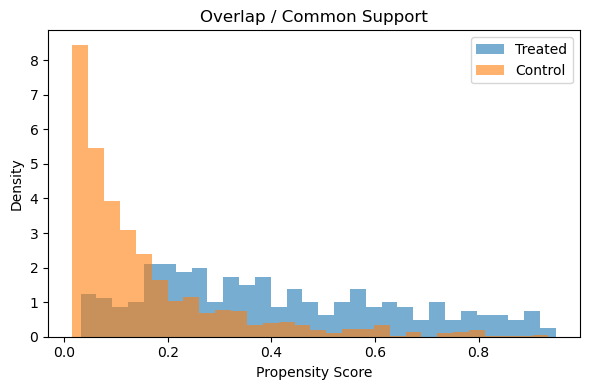}
    \caption{Propensity score overlap between treated and control groups. Common support spans [0.032, 0.935]; trimming 8.6\% (concentrated among controls) leaves overlap for estimation.}
    \label{fig:overlap}
\end{figure}

Table~\ref{tab:positivity_diagnostics} reports the overlap, effective-sample-size, and balance diagnostics used by the main positivity check; Figure~\ref{fig:overlap} shows the propensity-score overlap.

\begin{table}[t]
\centering
\caption{Overlap and balance diagnostics before and after trimming. Trimming reduces sample size modestly, narrows common support to regions shared by treated and control users, and keeps post-IPW covariate balance near the conventional 0.1 threshold (the largest absolute SMD after trimming is 0.104, for \texttt{Web3}). Overall effective sample sizes use stabilized weights; the treated and control rows are invariant to that choice. What to notice is that positivity is imperfect but remains adequate for the main analysis.}
\label{tab:positivity_diagnostics}
\begin{tabular}{lcc}
\toprule
Metric & Before Trim & After Trim \\
\midrule
\multicolumn{3}{l}{\textit{Sample Size}} \\
\quad Total N & 1,280 & 1,170 \\
\quad Treated & 263 & 261 \\
\quad Control & 1,017 & 909 \\
\midrule
\multicolumn{3}{l}{\textit{Propensity Score Range}} \\
\quad Treated & [0.032, 0.949] & [0.032, 0.910] \\
\quad Control & [0.016, 0.935] & [0.032, 0.935] \\
\quad Common support & [0.032, 0.935] & [0.032, 0.910] \\
\midrule
\multicolumn{3}{l}{\textit{Effective Sample Size (IPW)}} \\
\quad Overall & 836.0 & 741.6 \\
\quad Treated & 116.2 & 115.8 \\
\quad Control & 734.7 & 647.3 \\
\midrule
\multicolumn{3}{l}{\textit{Covariate Balance (|SMD| post-IPW)}} \\
\quad Total\_PvE\_Battle & 0.093 & 0.004 \\
\quad Total\_Session & 0.102 & 0.023 \\
\quad Web3 & 0.012 & 0.104 \\
\quad Time\_Play\_Level1 & 0.008 & 0.011 \\
\quad Max |SMD| post-IPW & 0.102 & 0.104 \\
\quad Max |SMD| before IPW & 1.16 & 1.06 \\
\bottomrule
\end{tabular}
\end{table}
\FloatBarrier

\section{Validation protocol}
\label{app:protocol}

Inputs: Observational telemetry with temporal slicing; minimal domain constraints; a target query \((T,Y)\).

Step 1 --- Data Preparation: Define unit of analysis and temporal windows; enforce temporal precedence; distinguish primitive from composite variables.

Step 2 --- Causal Discovery: Apply multiple algorithms under domain constraints; allow multiple admissible structures.

Step 3 --- Admissibility Gate: Proceed only with graphs satisfying: acyclicity, constraint compliance, valid identification strategy, adequate positivity.

Step 4 --- Effect Estimation: Estimate target effect under each admissible graph; record magnitude and uncertainty.

Step 5 --- Stability Analysis: Compare effects across algorithms; assess sensitivity to $\alpha$ thresholds; identify invariant vs. sensitive estimates.

Step 6 --- Falsification: Placebo intervention; data subsampling tests.

\textbf{Decision:}
\begin{itemize}
    \item Trust: Effect stable and falsification passed
    \item Caution: Sign-consistent but magnitude sensitive
    \item Reject: Effect unstable or falsified
\end{itemize}
The telemetry result falls under Caution: it is sign-consistent under every check but its magnitude depends on the subsample size and on the adjustment set (Section~\ref{sec:adjset}).

\section{Claim--assumption matrix}
We summarize the claims made in the paper (Table~\ref{tab:claim_assumption}), the assumptions required for each claim, how we test each claim, and what failure looks like.
\label{app:claims}

\begin{table}[!htbp]
\centering

\caption{Claim--assumption matrix. Each row states what is being evaluated, the required assumptions, how the paper tests it, and what failure looks like.}
\label{tab:claim_assumption}
\begin{tabularx}{\linewidth}{@{}>{\raggedright\arraybackslash}p{2.1cm}>{\raggedright\arraybackslash}X>{\raggedright\arraybackslash}X>{\raggedright\arraybackslash}X@{}}
\toprule
Claim & Assumption & Evidence & Failure condition \\
\midrule
Backdoor identification & Temporal order; no unblocked hidden path after adjustment & Constraint table; DoWhy ID check & No valid adjustment set exists \\
\addlinespace
Positivity satisfied & Common support of propensity scores & Overlap plot; standardized mean difference (SMD); effective sample size (ESS) & Effect undefined or driven by extrapolation \\
\addlinespace
Artifact check & No residual confounding or model misspecification & Placebo test ($p = 0.96$); subsampling sign stability & Placebo non-null; subsample sign flip \\
\addlinespace
Hidden-bias sensitivity & Unmeasured confounding weaker than the E-value & E-value ($= 1.66$, bound $1.48$), a sensitivity summary, not exclusion of confounding & A confounder with risk ratio $\geq 1.66$ to both $T$ and $Y$ would explain the effect away; moderate confounding cannot be ruled out \\
\addlinespace
Cross-algorithm agreement & Graphs differ only where the estimand does not depend on them & Constraint-based (PC) and score-based (GRaSP, BOSS) searches; graph comparison (Table~\ref{tab:graph_comparison}) & Agreement is on one adjustment set, not independent recovery; a different adjustment set or constraint specification gives a different estimate \\
\addlinespace
PAG caution & Partial orientation reflects latent-aware caution & PAG appendix; separate reporting & PAG and DAG agree only when query resolved \\
\bottomrule
\end{tabularx}
\end{table}
\FloatBarrier

\section{Run-level outcome taxonomy}
\label{app:taxonomy}

\begin{table}[!htbp]
\centering
\caption{Classification of the 21 DAG-returning discovery runs (PC, GRaSP, BOSS; 7 thresholds each) under full constraints (C3), scoped to the main \emph{backdoor} admissibility gate. A \emph{run} is one (algorithm, $\alpha$) tuple. Positivity is assessed at dataset level, and no run fails this gate after passing identification (0\% positivity failures); no admissible run is unstable (0\%), since all return the same estimate. PAG-returning methods (FCI, CFCI, FCI-MAX, GFCI, BOSS-FCI, GRaSP-FCI; 42 runs) are not evaluated under the DAG backdoor gate (Section~\ref{sec:admissibility}) and are reported separately in Appendix~\ref{app:pag}.}
\label{tab:taxonomy}
\begin{tabular}{@{}p{3.6cm}p{2.2cm}p{3.6cm}@{}}
\toprule
Bin & Fraction & Algorithms \\
\midrule
Admissible --- convergent & 100\% (21/21) & PC, GRaSP, BOSS (all $\alpha$) \\
\addlinespace
Inadmissible --- identification failure & 0\% & None at C3 (21 of 21 at C1; 11 of 21 at C0) \\
\bottomrule
\end{tabular}
\end{table}

\clearpage
\section{Admissible graph comparison}
\label{app:graphcomp}

\begin{figure}[!htbp]
\centering
\begin{tikzpicture}[
  n/.style={draw, rounded corners, minimum width=0.95cm, minimum height=0.5cm, font=\footnotesize},
  sh/.style={->, gray!70},
  ca/.style={->, thick, red!70!black},
  cb/.style={->, thick, blue!70!black},
  cc/.style={->, thick, green!50!black},
  >=Stealth
]
\node[n] (web3) at (0,1.5) {Web3};
\node[n] (pvp) at (2.4,3.0) {PvP};
\node[n] (tp) at (2.4,0) {TP1};
\node[n] (pve) at (5.0,2.3) {PvE};
\node[n] (ses) at (5.0,0.7) {Ses};
\node[n] (r1) at (7.4,1.5) {R1};
\draw[sh] (web3) -- (pvp);
\draw[sh] (web3) -- (tp);
\draw[sh] (web3) -- (ses);
\draw[sh] (web3) -- (r1);
\draw[sh] (pvp) -- (pve);
\draw[sh] (pvp) -- (ses);
\draw[sh] (pve) -- (r1);
\draw[sh] (pve) -- (ses);
\draw[sh] (ses) -- (r1);
\draw[ca] (tp) to[bend right=28] (r1);
\draw[ca] (pve) to[bend left=18] (tp);
\draw[cb] (tp) to[bend left=18] (pve);
\draw[cc] (web3) to[bend left=22] (pve);
\end{tikzpicture}
\caption{The three admissible graphs for the telemetry query under C2 and C3. Grey edges are shared by all three. Red edges appear only in graph A (\texttt{Total\_PvE\_Battle}$\to$\texttt{Time\_Play\_Level1} and \texttt{Time\_Play\_Level1}$\to$R1), blue in graphs B and C (\texttt{Time\_Play\_Level1}$\to$\texttt{Total\_PvE\_Battle}, reversed), green only in graph C (\texttt{Web3}$\to$\texttt{Total\_PvE\_Battle}). TP1 = \texttt{Time\_Play\_Level1}, PvE = \texttt{Total\_PvE\_Battle}, Ses = \texttt{Total\_Session}. In all three, parents(PvP) $=\{\texttt{Web3}\}$.}
\label{fig:admissible_graphs}
\end{figure}

\begin{table}[!htbp]
\centering
\caption{Admissible DAG runs (a converged run with a defined estimate, pooled over the constraint levels shown), their structural distinctness (pairwise SHD, a reversed edge counting as two errors), and the adjustment sets they imply. The 42 telemetry records at C2--C3 are the same 21 PC, GRaSP and BOSS runs at each level. On telemetry, estimates are identical within an adjustment set and differ across sets; on WoWAH this holds for the estimates within the risk-difference bounds $[-1,1]$, while the other admissible runs return out-of-bounds values.}
\label{tab:graph_comparison}
\setlength{\tabcolsep}{3pt}
\begin{tabularx}{\textwidth}{@{}>{\raggedright\arraybackslash}X>{\centering\arraybackslash}p{2.35cm}>{\centering\arraybackslash}p{2.2cm}>{\centering\arraybackslash}p{1.9cm}>{\centering\arraybackslash}p{3.25cm}@{}}
\toprule
Cohort & Admissible runs & Distinct graphs & Pairwise SHD & Distinct adjustment sets \\
\midrule
Telemetry, C2--C3 & 42 & 3 & 1--4 & 1 (\{\texttt{Web3}\}) \\
Telemetry, C0--C3 & 52 & 5 & 1--9 & 2 (\{\}, \{\texttt{Web3}\}) \\
WoWAH 336h, C0--C3 & 46 & 7 & 1--7 & 2 \\
WoWAH 168h, C0--C3 & 37 & 6 & 1--9 & 3 \\
\bottomrule
\end{tabularx}
\end{table}

\section{Leave-one-constraint-out}
\label{app:loo}

\begin{table}[!htbp]
\centering
\caption{Leave-one-constraint-out over 21 DAG-returning runs (PC, GRaSP and BOSS; 7 thresholds each). The WoWAH columns separate graph-identified runs from those whose estimates also lie within the risk-difference bounds $[-1,1]$. BOSS iteration-cap runs remain in the denominator as nonconverged. On telemetry the runs admitted without the platform constraint return the unadjusted difference 0.239 instead of 0.174. Estimator seeded. GRaSP is stochastic in the platform-dropped telemetry variant: over 20 repeats the admissible share ranges from 33.3 to 52.4 (mean 41.7), against the 47.6; the measurement-dropped value (0) was identical in all 140 GRaSP repeats.}
\label{tab:loo}
\begin{tabular}{@{}lccc@{}}
\toprule
Constraint dropped & Telemetry identified & WoWAH identified & WoWAH within bounds \\
\midrule
none (full set) & 100 & 66.7 & 33.3 \\
semantic & 100 & 66.7 & 33.3 \\
temporal & 66.7 & 33.3 & 0.0 \\
platform & 47.6 & 4.8 & 0.0 \\
measurement & 0.0 & 81.0 & 66.7 \\
\bottomrule
\end{tabular}
\end{table}

\section{Known-truth evaluation details}
\label{app:knowntruth}

Each sample is generated by a temporally ordered structural causal model over six variables (\texttt{Web3}, a baseline skill variable, PvP, \texttt{Total\_PvE\_Battle}, \texttt{Total\_Session}, R1) at $n=1{,}280$, and the interventional risk-difference ATE is computed exactly from the model. Discovery sees only the observed columns. The four regimes are: \emph{identifiable} (all confounders observed and measured before treatment, as declared for the telemetry case); \emph{latent confounder} (an unobserved common cause of PvP and R1); \emph{orientation ambiguous} (a nonlinear, observed baseline confounder whose linear association with treatment is near zero, so the edge is easily missed); and \emph{null} (no direct or mediated effect of PvP on R1). Each regime has 100 replications; PC, BOSS, GRaSP and FCI are run at three significance thresholds, giving 4 regimes $\times$ 100 $\times$ 4 $\times$ 3 $=$ 4,800 runs, of which 900 per regime are DAG runs. The admissibility gate is the empirical pipeline's: any graph with an unresolved edge (\texttt{---}, \texttt{o->}, \texttt{<->}, \texttt{o-o}) is rejected, and DoWhy identification supplies the adjustment set for the rest. Two things differ from the telemetry analysis: the estimator is a standardized logistic regression (with a squared skill term) rather than the seeded DRLearner, and the discovery constraints forbid PvP$\to$covariate edges, closer to the C4 set than to C3, with no measurement or semantic constraint. FCI outputs are PAGs, so they are not evaluated under the DAG gate and are summarized only through their latent-confounding warnings. All per-run outputs are released with the replication package.

On graph similarity, pooled across regimes SHD and absolute ATE error correlate ($+0.466$), but within regimes the correlation is weaker (between $+0.07$ and $+0.23$ in the three regimes with many surviving runs). SHD does not separate safe answers from unsafe ones: among low-SHD runs (SHD $\leq 4$, the median split) 51.1\% are still wrong by more than 5 pp, and perfect recovery, while safe, occurs in 8 of 2,615 admitted runs. Graph similarity is therefore a weak screen for decision usefulness rather than an uninformative one. Separately, the FCI latent-neighbourhood warning rate was only modestly higher in the latent-confounder regime (53.3\%) than in the identifiable regime (48.3\%), and a bidirected treatment--outcome mark was rare in both (3.3\% versus 1.7\%). This is setting-specific evidence (one DGP family, one sample size, one conditional-independence test), not a general conclusion about FCI. GRaSP is stochastic, so rare outcomes move between repetitions of the whole simulation: the safe share in the orientation-ambiguous regime was between 1.3\% and 4.6\% across repeated runs of the simulation, with the share wrong despite every check between 95.4\% and 98.6\%.

\section{Replication on WoWAH: cohort construction}
\label{app:wowah}

The replication uses the World of Warcraft Avatar History crawl~\citep{lee2011world}: 36.5M records over 91,064 avatars. Each avatar's column roles mirror the telemetry case: guild membership at $t_0$ (platform and social commitment, in the role of \texttt{Web3}), hours observed at level 1 (onboarding friction), samples in PvE zones before $t_0$ (baseline progression), and distinct sessions before $t_0$ (engagement intensity). The treatment is being seen in a PvP zone within the treatment window after $t_0$, and the outcome is being seen at all in the 24 hours that follow it. The cohorts of Table~\ref{tab:wowah} differ in the length of the treatment window (24, 168 and 336 hours). The PvP-eligibility landmark $t_0$ is the first time an avatar reaches level 10, derived from the observed progression data. Of the 91,064 avatars, 59,654 are first observed at level $\leq 1$; 58,655 of these are first seen after a 14-day burn-in; 11,795 reach PvP eligibility and are active until $t_0$; and requiring follow-up before the end of the crawl, at least 80\% crawler coverage of the outcome window, and a complete pre-$t_0$ covariate history leaves 9,592 (24h), 9,436 (168h) and 9,338 (336h) avatars.

\section{Reproducibility, impacts, and asset use}
\label{app:responsible_reporting}

\paragraph{Compute resources.}
All analyses were run on a single workstation-class CPU environment; the discovery and effect-estimation pipeline does not require GPU acceleration. The main experiments operate on 1,280 player-level observations with six analysis variables and complete in minutes on commodity hardware. Peak memory use is modest because the largest objects are tabular telemetry matrices, graph candidates, and bootstrap or subsampling summaries rather than large model checkpoints.

\paragraph{Broader impacts.}
The positive impact of the framework is more reliable causal decision support: it can help teams avoid acting on graph structures that are statistically plausible but do not identify the target effect. The same tools could also be misused to optimize engagement or retention without adequate user welfare, privacy, or fairness review. We therefore treat the method as a diagnostic layer for decision quality, not as an automatic policy engine, and recommend pairing effect-level validation with institutional review of data governance and downstream product interventions.

\paragraph{Reproducibility.}
The results under \texttt{experiment/} were produced with two software builds, both recorded in the package (\texttt{requirements.txt} for build B, \texttt{requirements-build-A.txt} for build A). Build A (Python 3.12.12; py-tetrad 0.1.2 at commit \texttt{38bfa024}, Tetrad jar of 2025-12-28; DoWhy 0.14, EconML 0.16.0, causal-learn 0.1.4.3, JPype1 1.6.0, scikit-learn 1.6.1, pandas 2.3.3, NumPy 1.26.4, statsmodels 0.14.5) ran PC, GRaSP and the six PAG-returning algorithms; it provides FCI-MAX, CFCI and BOSS-FCI, but its BOSS does not terminate under required edges. Build B (Python 3.13.9; py-tetrad at commit \texttt{fbf067e7}, jar of 2025-10-28; DoWhy 0.14, EconML 0.16.0, causal-learn 0.1.4.4, scikit-learn 1.6.1, pandas 2.3.3, NumPy 2.3.5, statsmodels 0.14.6) adds the BOSS iteration cap and ran BOSS, but has no FCI-MAX, CFCI or BOSS-FCI wrapper. Both used Java/JDK 25.0.1, and their estimates agree to machine precision wherever the same graph is returned. Re-running 428 tasks (every dataset and constraint level, two thresholds) reproduced the stored graph in 414; the 14 exceptions are 13 PAG-returning runs (FCI at $\alpha=0.001$ on WoWAH, GRaSP-FCI, BOSS-FCI) and one PC graph whose identifiability outcome is unchanged, so none affects the DAG gate. The doubly robust estimator is explicitly seeded (seed 20260725). Tetrad's seed parameter does not make GRaSP deterministic in either build, so GRaSP is stochastic where its search is under-constrained: telemetry C0, C1, C4, \textsc{LOO}-platform and \textsc{LOO}-measurement, and WoWAH 168h C0. We repeated those cells 20 times and report the spread; the repeats are released with the per-run outputs. In build A all 112 FCI-MAX runs across the four cohorts completed successfully (mean 5.9 s).

\paragraph{Data access, consent, and ethics review.}
The telemetry dataset is released as a de-identified, six-variable, user-level analytic subset (1,280 observations) under CC BY 4.0, without user identifiers, wallet addresses, device identifiers, IP addresses, raw timestamps, or raw event logs. Users consented to gameplay telemetry collection under the game's Terms of Service and/or Privacy Policy, and no prospective intervention, crowdsourcing task, or human-subject experiment was conducted. No institutional review (IRB) was sought, because the analysis uses only de-identified data collected under the game's terms of service.

\paragraph{Licenses and assets.}
The paper uses original aggregate figures and tables derived from the authors' analysis pipeline, together with standard open-source scientific Python and causal-discovery software cited in the main text and appendix. The replication code is released under the MIT License, and the analytic telemetry dataset is released separately under CC BY 4.0. The replication of Section~\ref{sec:wowah} uses the third-party World of Warcraft Avatar History crawl~\citep{lee2011world}, which we credit; its authors distribute it for free download through the ACM MMSys 2011 dataset track, whose page states no license or terms of use, and we do not redistribute it in the paper. The replication package documents the software dependencies and license information needed to reproduce the reported analyses. No third-party media, pretrained models, scraped public datasets, or proprietary benchmark assets are redistributed as part of the paper.
\end{document}